\def\year{2017}\relax
\documentclass[letterpaper]{article}
\usepackage{aaai17}
\usepackage[ruled,vlined]{algorithm2e}
\usepackage{amsmath}
\usepackage{blindtext}
\usepackage{times}
\usepackage{helvet}
\usepackage{graphicx}
\usepackage{courier}
\usepackage{amsmath}
\usepackage{color}
\DeclareMathOperator{\Tr}{Tr}
\begin{document}

%
\title{Learning Deep Latent Spaces for Multi-Label Classification}
\author{Chih-Kuan Yeh$^1$\thanks{ - indicates equal contribution.}, Wei-Chieh Wu$^2$\footnotemark[1], Wei-Jen Ko$^2$, Yu-Chiang Frank Wang$^1$ \\
	$^1$Research Center for IT Innovation, Academia Sinica, Taipei, Taiwan\\
	$^2$Department of Electrical Engineering, National Taiwan University, Taipei, Taiwan\\
	jason6582@gmail.com, \{b01504088, b01901162\}@ntu.edu.tw, ycwang@citi.sinica.edu.tw
}

\maketitle
\begin{abstract}
Multi-label classification is a practical yet challenging task in machine learning related fields, since it requires the prediction of more than one label category for each input instance. We propose a novel deep neural networks (DNN) based model, \textbf{C}anonical \textbf{C}orrelated \textbf{A}uto\textbf{E}ncoder (C2AE), for solving this task. Aiming at better relating feature and label domain data for improved classification, we uniquely perform joint feature and label embedding by deriving a deep latent space, followed by the introduction of label-correlation sensitive loss function for recovering the predicted label outputs. Our C2AE is achieved by integrating the DNN architectures of canonical correlation analysis and autoencoder, which allows end-to-end learning and prediction with the ability to exploit label dependency. Moreover, our C2AE can be easily extended to address the learning problem with missing labels. Our experiments on multiple datasets with different scales confirm the effectiveness and robustness of our proposed method, which is shown to perform favorably against state-of-the-art methods for multi-label classification.


\end{abstract}

\section{Introduction}
\label{sec:intro}

With rich information presented in multimedia data, many real-world classification tasks require one to assign more than one label to each instance. For example, multiple types of objects in an image need to be annotated, or different identities need to be determined from an audio clip~\cite{zhang2014review}. Thus, different from standard multi-class recognition problems (i.e., only one class label for each input data), multi-label classification typically requires additional efforts in extracting and describing the associated data/label information to produce satisfactory performances.

By dividing the original multi-label classification problem into multiple independent binary classification tasks, binary relevance~\cite{tsoumakas2006multi} is a straightforward technique and solution, which has been widely applied by users in related fields. However, in addition to the concern of high computational costs, such techniques cannot identify the correlation between label information, which would limit the resulting prediction performance. As a result, methods proposed by~\cite{read2011classifier,cheng2010bayes} aim at exploiting the cross-label dependency by assuming label prior information. Unfortunately, since these approaches perform a series of classification for multi-label prediction, parallel implementation is not applicable if reducing computation loads is desirable.

Deriving a latent label space with reduced dimensionality is also a popular technique for multi-label classification~\cite{balasubramanian2012landmark,tai2012multilabel,chen2012feature,bi2013efficient,hsu2009multi,zhang2012maximum,zhou2012compressed,lin2014multi,yu2014large,li2015multi,bhatia2015sparse}. Its goal is to transform the label space into a latent subspace, followed by the association between the projected input and label data for classification purposes. With a proper decoding process which maps the projected data back to the original label space, the task of multi-label prediction is achieved. Since the learning of such latent subspaces not only reduces the classification time, the correlation between the labels can be implicitly exploited. Instead of observing latent spaces with reduced dimensions, ~\cite{tsoumakas2011random,ferng2013multilabel} proposed to derive high-dimensional label embedding space for performing the above task. Nevertheless, the above latent space learning algorithms can all be viewed as label embedding based approaches. Moreover, the ability to handle missing labels during the learning of multi-label classification models is also practical for real-world application like image annotation. Incomplete labeled data during training might result in noisy classifiers with insufficient prediction capability. While this is typically not well addressed in existing methods, \cite{wu2014multi} chose a transductive setting with label smoothness regularization, and \cite{wu2015ml} approached the problem by formulating a convex quadratic matrix optimization problem.

Among the first to utilize neural network architectures, BP-MLL~\cite{zhang2006multilabel} not only treated each output node as a binary classification task, and relied on the architecture itself to exploit the dependency across labels. Later, it was extended by~\cite{nam2014large} with additional deep neural networks (DNN) techniques. Some recent works proposed different loss functions \cite{gong2013deep} or architectures \cite{wei2014cnn} for further improving the performance. For example, CNN-RNN \cite{wang2016cnn} chose to learn a linear label embedding function, with label co-occurrence information observed by recurrent neural networks (RNN). However, since only linear embedding was considered, higher order dependency between different labels might not be successfully discovered.

In this paper, we present a novel DNN-based framework, Canonical-Correlated Autoencoder (C2AE), for multi-label classification. Different from most label embedding based methods which typically view label embedding and prediction as two separate tasks, our C2AE advances deep canonical correlation analysis (DCCA) and autoencoder to learn a feature-aware latent subspace for label embedding and multi-label classification. Moreover, with label-correlation aware loss functions introduced at the decoding outputs, our C2AE is able to better exploit cross-label dependency during both label embedding and prediction processes. The main contributions of this paper are highlighted as follows:

\begin{itemize}
	\item By utilizing and integrating the architectures of deep canonical correlation analysis and autoencoder, our Canonical-Correlated Autoencoder (C2AE) is among the first DNN-based label embedding frameworks for multi-label classification.
	\item Our C2AE is able to perform feature-aware label embedding and label-correlation aware prediction. The former is realized by joint learning of DCCA and the encoding stage of autoencoder, while the latter is achieved by the introduced loss functions for the decoding outputs.
	\item Without modifying the proposed architecture, our C2AE can be easily extended to handle missing label problems. Our experiments verify that we perform significantly better than state-of-the-art approaches on multi-label classification tasks with/without missing labels.	
\end{itemize}

\section{Related Work}
\label{sec:rewo}

While binary relevance~\cite{tsoumakas2006multi} is among the popular techniques for multi-label classification, the lack of sufficient ability to discover interdependency between labels would be its concern. 

To address the above issue, approaches based on classifier chains were proposed. For example, probabilistic classifier chains (PCC) aim at capturing conditional label
dependency via the product rule of probabilities \cite{cheng2010bayes}. While beam search \cite{kumar2013beam} and advanced inference procedure \cite{dembczynski2012analysis} were further extended from PCC, these approaches are typically computationally expensive, and cannot be easily extended to problems with a large number of labels.

Label embedding (LE) is another popular strategy for multi-label classification. It transforms the label vectors into a subspace with latent embedding of the corresponding information, and the correlation between labels can be implicitly described. With additional mapping (from the input vectors) and decoding (for prediction) stages derived for this latent label space, one can perform multi-label prediction with reduced computation costs~ \cite{hsu2009multi,balasubramanian2012landmark,tai2012multilabel,chen2012feature,zhang2012maximum,zhou2012compressed,bi2013efficient,yu2014large,lin2014multi,li2015multi}. For example, in \cite{hsu2009multi}, the label embedding was obtained via random projections, while principal component based projections (e.g., principal label space transformation (PLST)~\cite{tai2012multilabel} and its conditional version (CPLST)~\cite{chen2012feature}) were later utilized. Variants of LE like~\cite{lin2014multi}, ~\cite{bhatia2015sparse}, and~\cite{li2015multi} are also available, which aim at improving the predictability and recoverability of proposed models. Recently, ~\cite{wang2016cnn} presented CNN-RNN, which applied linear label embedding followed by recurrent neural networks (RNN) for better identifying the co-occurrence of labels.

We note that, existing LE approaches typically consider linear embedding functions, while some apply standard kernel functions (e.g., low-degree polynomial kernels) for nonlinear embedding. Moreover, only few methods jointly utilize the input feature space for label embedding (e.g.,~\cite{chen2012feature,lin2014multi,li2015multi}). In this paper, we advance deep neural networks for exploiting label correlation during the embedding process. In particular, we propose Canonical-Correlated Autoencoder (C2AE), which can be viewed as a feature-aware label embedding framework with ability in exploiting label interdependency during both embedding and prediction processes. We will detail our proposed DNN model in the following section.

\begin{figure}[t!]
	\centering
	\includegraphics[width=0.48\textwidth]{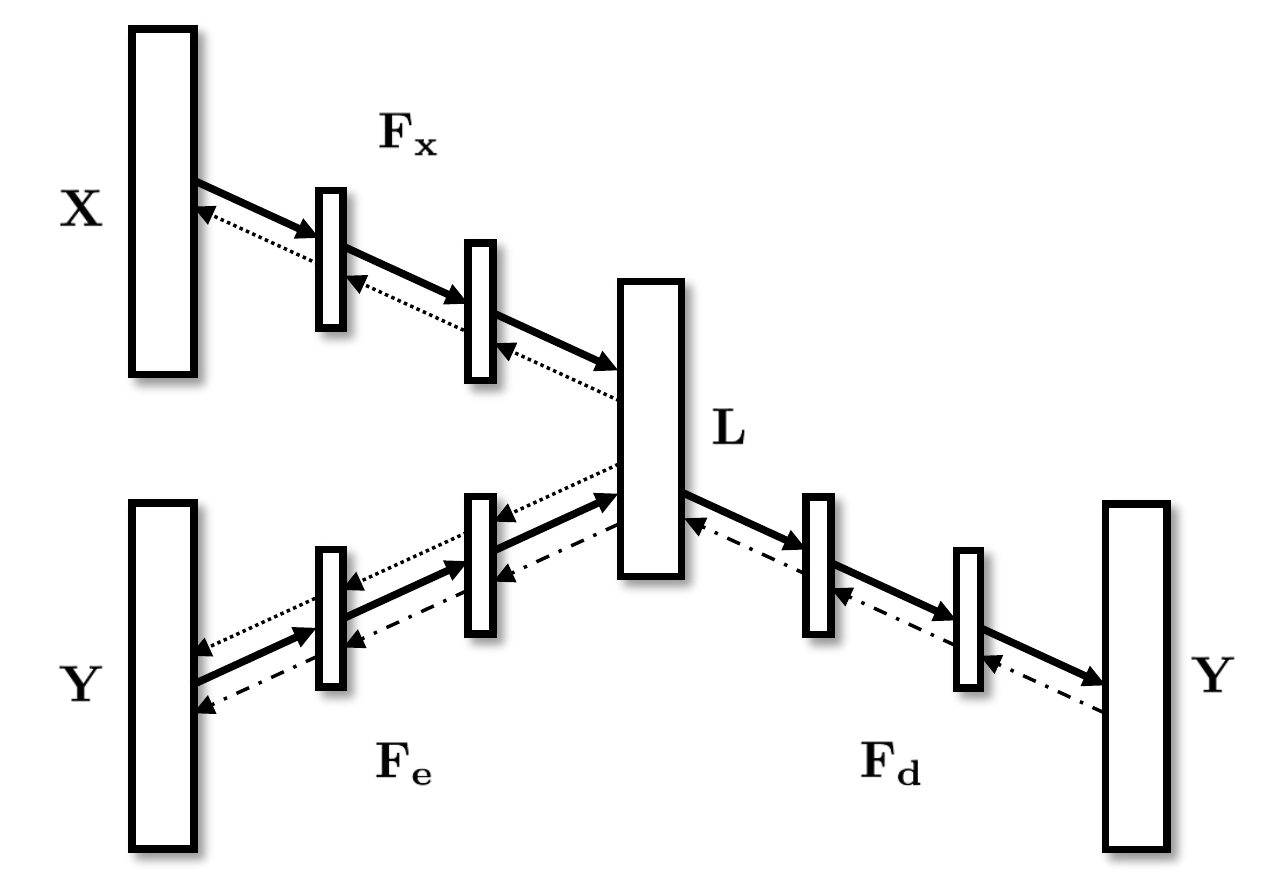}
	\vspace{-6mm}
	\caption{\small The proposed architecture of Canonical-Correlated Autoencoder (C2AE), which learns a latent space $L$ via nonlinear mappings of $\textbf{F}_x$, $\textbf{F}_e$, and $\textbf{F}_d$. Note that $\textbf{X}$ and $\textbf{Y}$ are the input and label data, respectively.}
	\vspace{0mm}
	\label{fig:overview}
\end{figure}

\section{Our Proposed Method}
\subsection{Canonical-Correlated Autoencoder (C2AE)}
Let $D=\{(\textbf{x}_i,\textbf{y}_i)\}_{i=1}^N = \{\textbf{X},\textbf{Y}\}$ denote a set of $d$ dimensional training instances $\textbf{X} \in {\rm I\!R}^{d\times N}$ and the associated labels $\textbf{Y} \in {\{0,1\}}^{m\times N}$, where $N$ and $m$ are the numbers of instances and label attributes, respectively. By observing $D$, the goal of multi-label classification is to derive a proper learning model, so that the label $\hat{\textbf{y}}$ of a test instance $\hat{\textbf{x}}$ can be predicted accordingly.

Motivated by label embedding and the recent developments in deep learning, we propose a novel DNN architecture of Canonical-Correlated Autoencoder (C2AE), as depicted in Figure~\ref{fig:overview}. Our C2AE utilizes Deep Canonical Correlation Analysis (DCCA) and autoencoder structures, which learns a latent subspace from both feature and label domains for multi-label classification.

As illustrated in Figure~\ref{fig:overview}, our C2AE (denoted by $\mathbf{\Theta}$) integrates two effective DNN models (i.e., DCCA and autoencoder) with three mapping functions to be determined: feature mapping $\textbf{F}_x$, encoding function $\textbf{F}_e$, and decoding function $\textbf{F}_d$. During the training stage, the input of C2AE are the observed training instances $\textbf{X}$ and their labels $\textbf{Y}$, while the recovered output is the label of interest $\textbf{Y}$ (i.e., same as the input labels). Aiming at determining the latent space $\textbf{L}$, the DCCA component of our C2AE associates $\textbf{X}$ and $\textbf{Y}$, while the autoencoder part enforces the output is recovered as $\textbf{Y}$. Thus, the objective function of C2AE can be formulated as follows:
\begin{equation}\label{alg:obj}
\mathbf{\Theta} = \min_{\mathbf{F}_x,\mathbf{F}_e,\mathbf{F}_d} \mathbf{\Phi}(\mathbf{F}_x,\mathbf{F}_e) + \alpha \mathbf{\Gamma}(\mathbf{F}_e,\mathbf{F}_d),
\end{equation}
where $\Phi(\mathbf{F}_x,\mathbf{F}_e)$ and $\Gamma(\mathbf{F}_e,\mathbf{F}_d)$ denote the losses at the latent space and output of C2AE, respectively. And, we have the parameter $\alpha$ balances between the above two types of loss functions.

Once the learning of our C2AE is complete, it can be easily applied for predicting the labels of test inputs. To be more precise, a test input $\hat{\mathbf{x}}$ will be first transformed into the derived latent space by $\mathbf{F}_x$, followed by the decoding mapping of $\mathbf{F}_d$ for predicting its output label $\hat{\mathbf{y}}$ (i.e., $\hat{\mathbf{y}} = \mathbf{F}_d(\mathbf{F}_x(\hat{\mathbf{x}}))$.

\subsubsection{Learning Deep Latent Spaces for Joint Feature \& Label Embedding}$\\$

We now discuss why we advance DCCA in our C2AE for feature and label-aware embedding. For the sake of completeness, we first briefly review the ideas of CCA and DCCA~\cite{hotelling1936relations,andrew2013deep,wang2015deep}.

As a standard statistical technique for relating cross-domain data (e.g., input feature data $\mathbf{X}$ and their label data $\mathbf{Y}$), CCA determines linear projection matrices $\mathbf{W}_1$ and $\mathbf{W}_2$ for each domain, aiming at observing a subspace in which the correlation of projected data is maximized (i.e., ${corr}(\mathbf{W}_1^T\mathbf{X},\mathbf{W}_2^T\mathbf{Y})$). With the two linear projections replaced by DNNs, DCCA solves the same objective function with the DNN models learned/updated by gradient descent techniques~\cite{andrew2013deep}.

To determine $\mathbf{\Phi}(\mathbf{F}_x,\mathbf{F}_e)$ in \eqref{alg:obj}, we adapt the idea of \cite{kettenring1971canonical} and rewrite the correlation-based objective function as the following deep version:

\begin{equation}\label{eq:DCCA}
\begin{aligned}
& \underset{\mathbf{F}_x, \mathbf{F}_e}{\text{min}}
& & \|\mathbf{F}_x(\mathbf{X})-\mathbf{F}_e(\mathbf{Y})\|_F^2 \\
& \text{s.t.}
& & \mathbf{F}_x(\mathbf{X})\mathbf{F}_x(\mathbf{X})^T = \mathbf{F}_e(\mathbf{Y})\mathbf{F}_e(\mathbf{Y})^T = \mathbf{I},
\end{aligned}
\end{equation}
where $\mathbf{F}_x(\mathbf{X})$ and $\mathbf{F}_e(\mathbf{Y})$ denote the transformed feature and label data in the derived latent space $\mathbf{L}$, respectively. And, $\mathbf{I} \in {\rm I\!R}^{l\times l}$ is the identity matrix, where $l$ is the dimension of the latent space $L$. As explained in~\cite{kettenring1971canonical}, the above identity constraint would make the above formulation equivalent to the standard CCA objection function of correlation maximization. Compared to the standard CCA optimization task, the above formulation allows us to calculate the network loss and the corresponding gradient descent function efficiently.



By solving $\mathbf{F}_x(\mathbf{X})$ and $\mathbf{F}_e(\mathbf{Y})$ in \eqref{eq:DCCA} with DNN models, we enforce the learned deep latent space to jointly associate feature and label data. It is worth noting that, while existing multi-label classification approaches based on label embedding \cite{tai2012multilabel,chen2012feature,lin2014multi,li2015multi} perform subspace learning using feature and/or label data, they typically learn an additional model relating feature data and the derived subspace for prediction purposes. In other words, the tasks of label embedding and multi-label prediction are performed separately, which might not be preferable. In our work, we not only utilize \eqref{eq:DCCA} for joint feature and label embedding with classification guarantees, our integration with autoencoder architectures further allows satisfactory recoverability for prediction purposes (see the following subsection for details).




\subsubsection{Learning and Recovering Label-Correlated Outputs}$\\$

With the DCCA component in our C2AE performing DCCA for joint feature and label embedding, we further advance the autoencoder in C2AE for recovering label outputs, with a particular goal of preserving cross-label dependency.

Inspired by \cite{zhang2006multilabel}, we introduce a label-correlation aware loss function at the output of our C2AE, which is determined as follows:
\small
\begin{equation} \label{eq:AE}
\begin{aligned}
&\Gamma(\mathbf{F}_e,\mathbf{F}_d) = \sum_{i=1}^{N} E_i \\
E_i = \frac{1}{|\mathbf{y}_i^1||{\mathbf{y}_i^0}|}
& \sum_{(p,q)\in \mathbf{y}_i^1 \times {\mathbf{y}_i^0}}\exp(\mathbf{F}_d(\mathbf{F}_e(\mathbf{y}_i))^q - \mathbf{F}_d(\mathbf{F}_e(\mathbf{y}_i))^p),
\end{aligned}
\end{equation}
\normalsize
where $\mathbf{y}_i^1$ denotes the set of the positive labels in $\mathbf{y}_i$ for the $i$th instance $\mathbf{x}_i$, and $\mathbf{y}_i^0$ is that of the negative labels. Given the input $\mathbf{x}_i$, $\mathbf{F}_d(\mathbf{F}_e(\mathbf{x_i}))^p$ returns the $p$th entry of the C2AE output. Thus, minimizing the above loss function is equivalent to maximizing the prediction outputs of all positive-negative label attribute pairs, which implicitly enforces the preservation of label co-occurrence information. If standard mean square error or cross-entropy losses are considered, such label dependency cannot be successfully identified.

With the above loss function, our C2AE integrating DCCA and autoencoder can be viewed as an end-to-end DNN, which performs joint feature/label embedding and label-correlate aware prediction in a unified model. To be more precise, we are able to learn feature embedding $\textbf{F}_x$, label embedding $\textbf{F}_e$, and label prediction $\textbf{F}_d$ in a unified framework. As noted earlier, most existing linear or nonlinear label-embedding based approaches derive the above models separately with no performance correlation guarantees. Later in our experiments, we will verify the effectiveness of our approach over such methods.

\subsubsection{Learning from Data with Missing Labels}$\\$

As highlighted earlier, our C2AE can be further extended to multi-label classification problems with missing labels. That is, we need to learn a robust C2AE model, when missing labels during the training stage are expected.

To solve this challenging yet practical task, we now easily apply a more general setting for determining the loss function for our C2AE. More specifically, for an instance with positive, negative, and some missing label attributes, we determine the loss function of \eqref{eq:AE} by calculating the losses derived from known label pairs only (i.e, available positive-negative label pairs). This would make our C2AE robust to missing labels, and exhibits sufficient abilities in exploiting the label dependency from the known label attributes. 

In addition to extending our loss function at the output layer of C2AE for handling data with missing labels, we also perform a simple preprocessing stage for such data before feeding them into our network. To be more precise, we set the positive labels in an instance to be 1, the missing labels to be 0, and the negative labels to be $-\frac{|\mathbf{y}_i^1|}{|\mathbf{y}_i^0|}$ for keeping the average of the labels to 0. This is to guarantee that the missing labels would not be fed into the DNN model since its value is set to 0, which effectively suppresses the noise (coming from the missing labels) to be mapping into the latent space.



\subsection{Optimization}

To learn the model of C2AE, we need to solve the optimization problem of \eqref{alg:obj}, in which the loss terms $\Phi(\mathbf{F}_x,\mathbf{F}_e)$ and $\Gamma(\mathbf{F}_e,\mathbf{F}_d)$ are calculated at the latent space and the output of C2AE, respectively.

Similar to the derivation of existing DNN models, we apply the technique of gradient descent for each loss term for updating the corresponding network parameters. As shown in Figure \ref{fig:overview}, the gradient of $\mathbf{\Phi}(\mathbf{F}_x,\mathbf{F}_e)$ updates the feature mapping $\mathbf{F}_x$ and encoding $\mathbf{F}_e$, while that of $\mathbf{\Gamma}(\mathbf{F}_e,\mathbf{F}_d)$ updates both encoding $\mathbf{F}_e$ and decoding functions $\mathbf{F}_d$.

To calculate the gradient term of $\mathbf{\Phi}(\mathbf{F}_x,\mathbf{F}_e)$, we reformulate \eqref{eq:DCCA} with the aid of Lagrange multipliers:

\begin{equation*}
{\Phi}(\mathbf{F}_x,\mathbf{F}_e)
= \Tr(\mathbf{C_1}^T\mathbf{C_1})
+ \lambda\Tr(\mathbf{C_2}^T\mathbf{C_2} + \mathbf{C_3}^T\mathbf{C_3}),
\end{equation*}
where
\begin{equation*}
\begin{aligned}
&\mathbf{C_1} = \mathbf{F}_x(\mathbf{X}) - \mathbf{F}_e(\mathbf{Y}) \\
&\mathbf{C_2} = \mathbf{F}_x(\mathbf{X})\mathbf{F}_x(\mathbf{X})^T - \mathbf{I} \\
&\mathbf{C_3} = \mathbf{F}_e(\mathbf{Y})\mathbf{F}_e(\mathbf{Y})^T - \mathbf{I}.
\end{aligned}
\end{equation*}

\noindent Thus, the gradient of ${\Phi}(\mathbf{F}_x,\mathbf{F}_e)$ with respect to $\mathbf{F}_x(\mathbf{X})$ and $\mathbf{F}_e(\mathbf{Y})$ can be derived as:

\begin{equation}\label{eq:phi1}
\frac{\partial{\Phi}(\mathbf{F}_x,\mathbf{F}_e)}{\partial \mathbf{F}_x(\mathbf{X})}
= 2\mathbf{C_1}
+ 4\lambda\mathbf{F}_x(\mathbf{X})\mathbf{C_2},
\end{equation}

\begin{equation}\label{eq:phi2}
\frac{\partial{\Phi}(\mathbf{F}_x,\mathbf{F}_e)}{\partial \mathbf{F}_e(\mathbf{Y})}
= 2\mathbf{C_1}
+ 4\lambda\mathbf{F}_e(\mathbf{X})\mathbf{C_3},
\end{equation}



Next, we discuss how to calculate the gradient of $\Gamma(\mathbf{F}_e,\mathbf{F}_d)$ (as determined in \eqref{eq:AE}) with respect to each $\mathbf{F}_d(\mathbf{F}_e(\mathbf{x_i}))^j$. For simplicity, we let $\mathbf{c}_i^j = \mathbf{F}_d(\mathbf{F}_e(\mathbf{y_i}))^j$, and thus the above gradient can be derived as follows:

\begin{equation*}\label{eq:gamma1}
\frac{\partial {\Gamma}(\mathbf{F}_e,\mathbf{F}_d)}{\partial \mathbf{c}_i^j} = \sum_{i=1}^{N} \frac{\partial {E}_i}{\partial \mathbf{c}_i^j}
\end{equation*}
\begin{equation}\label{eq:gamma2}
\frac{\partial {E}_i}{\partial \mathbf{c}_i^j} = \left\{
\begin{aligned}
{-\frac{1}{|\mathbf{y}_i^1||{\mathbf{y}_i^0}|} \sum_{q\in {\mathbf{y}_i^0}} \exp(-(\mathbf{c}_i^j - \mathbf{c}_i^q))} \text{, if } j \in \mathbf{y}_i^1 \\
{\frac{1}{|\mathbf{y}_i^1||{\mathbf{y}_i^0}|} \sum_{p\in{\mathbf{y}_i^1}} \exp(-(\mathbf{c}_i^p - \mathbf{c}_i^j))} \text{, if } j \in \mathbf{y}_i^0,
\end{aligned}
\right.
\end{equation}
where $\mathbf{y}_i^1$ denotes the set of the positive labels in $\mathbf{y}_i$ for the $i$th instance $\mathbf{x}_i$, and $\mathbf{y}_i^0$ is that of the negative labels.

With the above derivations, we can learn our C2AE by gradient descent, and the pseudo code is summarized in Algorithm \ref{alg:learning}. Once the learning of C2AE is complete, label prediction of a test input $\hat{\mathbf{x}}$ can be easily achieved by rounding $\hat{\mathbf{y}} = \mathbf{F}_d(\mathbf{F}_x(\hat{\mathbf{x}}))$.

\begin{algorithm}[t]
	\label{alg:learning}
	\DontPrintSemicolon
	\caption{Learning of C2AE}
	\KwIn{Feature matrix $\mathbf{X}$, label matrix $\mathbf{Y}$, parameter $\mathbf{\alpha}$, and dimension $l$ of the latent space}
	\KwOut{$\mathbf{F}_x$, $\mathbf{F}_d$, and $\mathbf{F}_e$}
	Randomly initialize $\mathbf{F}_x$, $\mathbf{F}_d$, $\mathbf{F}_e$.
	\\
	\Repeat{Converge}
	{
		Randomly select a batch of data $S[\mathbf{X}]$ and $S[\mathbf{Y}]$ \\
		Define the loss function by \eqref{alg:obj}\\
		Perform gradient descent on $\mathbf{F}_d$ by \eqref{eq:gamma2}\\
		Perform gradient descent on $\mathbf{F}_x$ by \eqref{eq:phi1}\\
		Perform gradient descent on $\mathbf{F}_e$ by \eqref{eq:phi2} and \eqref{eq:gamma2}\\
	}
\end{algorithm}

\section{Experiments}
\subsection{Datasets and Settings}
\begin{table}[]
	\centering
	\caption{Datasets considered for performance evaluation.}
	\label{my-label}
	\begin{tabular}{l|l|l|l|l}
		\hline
		dataset              & \#labels & \#instances & \begin{tabular}[c]{@{}l@{}}\#feature\end{tabular} & \begin{tabular}[c]{@{}l@{}}\#cardinality\end{tabular} \\ \hline
		\textit{iaprtc12}  & 291      & 19,627       & 1000                                                           & 5.7                                                                            \\
		\textit{mirflickr} & 38       & 25,000       & 1000                                                           & 4.7                                                                            \\
		\textit{espgame}   & 268      & 23,641       & 1000                                                           & 4.7
		\\
		\textit{tmc2007}   & 22       & 28,596     & 500                                                           & 2.1                                                                           \\
		\textit{NUS-WIDE}   & 81       & 269,648     & 4096                                                           & 1.9
		\\ \hline
	\end{tabular}
\end{table}

\begin{figure*}[t!]
	\centering
	\includegraphics[width=1.02\textwidth]{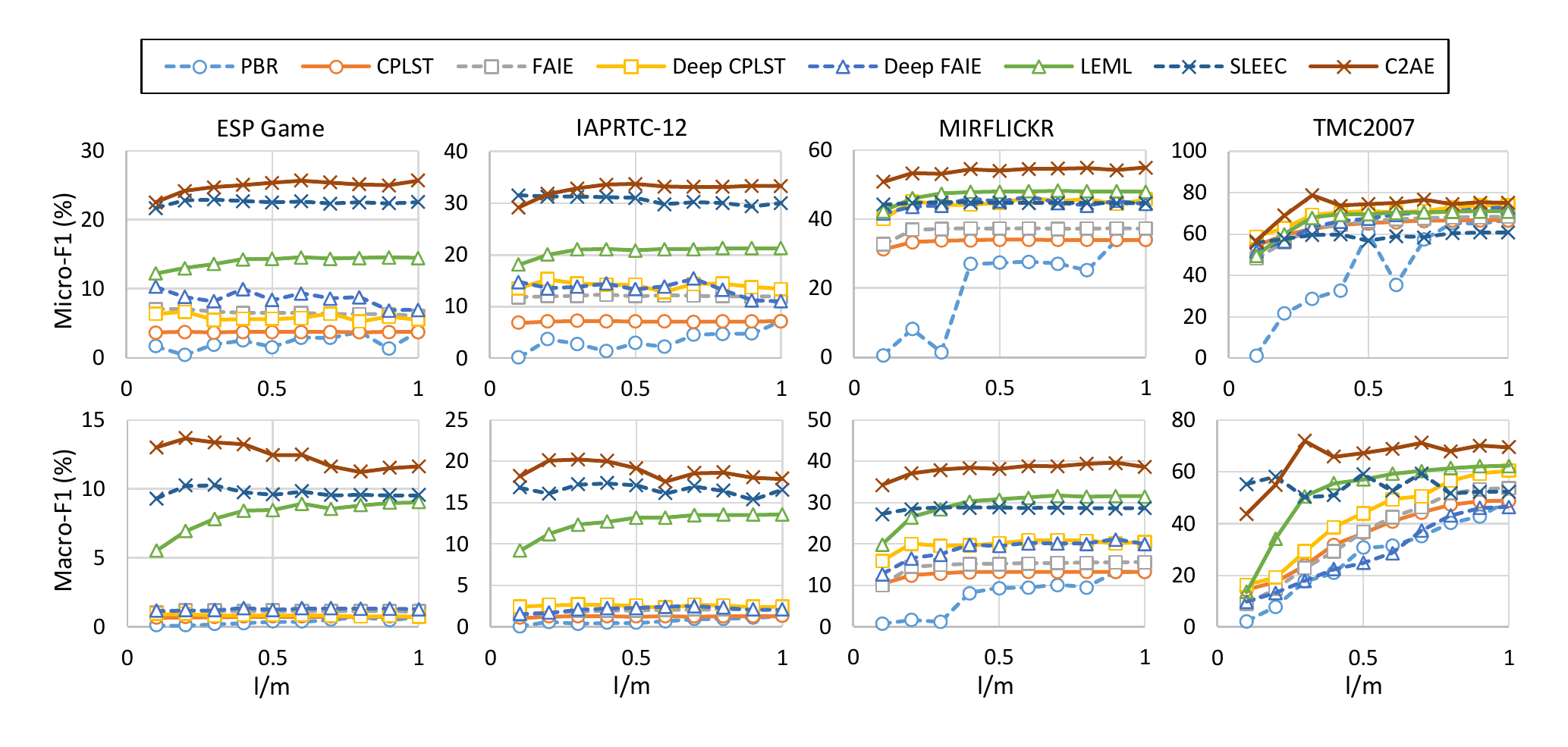}
	\vspace{-10mm}
	\caption{\small Performance comparisons in terms of Micro-F1 and Macro-F1 with different latent space dimension ratios ($l$/$m$).}
	\vspace{-0mm}
	\label{fig:LSDR}
\end{figure*}

To evaluate the performance of our proposed method, we consider the following datasets for experiments: \textit{iaprtc12}, \textit{ESPGame}, \textit{mirflickr}, \textit{tmc2007}, and \textit{NUS-WIDE}. The first three datasets are image datasets used in \cite{guillaumin2009tagprop}, where 1000-dimensional Bag-of-Words features (based on SIFT) are extracted. We note that \textit{tmc2007} is a large-scale text dataset downloaded from Mulan \cite{tsoumakas2011mulan}, and \textit{NUS-WIDE} is a large-scale image dataset typically applied for image annotation tasks. The details of each dataset are listed in Table \ref{my-label}. For \textit{NUS-WIDE}, we follow the setting of \cite{gong2013deep} by discarding the instances with no positive labels and randomly select 150,000 instances for training and the remaining for testing. For fair comparisons with other CNN-based methods, we extract 4096-dimensional fc-7 feature for NUS-WIDE using a pre-trained AlexNet model.\footnote{The experimental code implemented in matlab can be found at https://github.com/yankeesrules/C2AE.}

For the architecture of our C2AE, we have $F_x$ composed of 2 layers of fully connected layers, while $F_d$ and $F_e$ are both single fully connected layer structures. For each fully connected layer, a total of 512 neurons are deployed. A leaky ReLU activation function is considered, while the batch size is fixed as 500. To select the parameters for C2AE, we randomly hold 1/6 of our training data for validation (with $\alpha$ selected from [0.1, 10] and $\lambda$ fixed as 0.5). We also perform the same validation process for selecting the parameters (including the threshold for predicting the final labels) for other methods to be compared in our experiments. As for the evaluation metrics, we consider micro-F1 and macro-F1 \cite{tang2009large}.

\subsection{Comparisons with Label Embedding based Approaches}
We first consider the approaches based on label embedding for comparisons: Conditional Principal Label Space Transformation (CPLST) \cite{chen2012feature}, Feature-aware Implicit Label space Encoding (FaIE) \cite{lin2014multi}, Low rank Empirical risk minimization for Multi-Label Learning (LEML) \cite{yu2014large}, Sparse Local Embeddings for Extreme Multi-label Classification (SLEEC) \cite{bhatia2015sparse}, and the baseline method of partial binary relevance (PBR) \cite{chen2012feature}. In addition, we replace the linear regressors in CPLST and FAIE by DNN regressors, and denote such methods as Deep CPLST and Deep FAIE.

\begin{figure}[t!]
	\centering
	\includegraphics[width=0.49\textwidth]{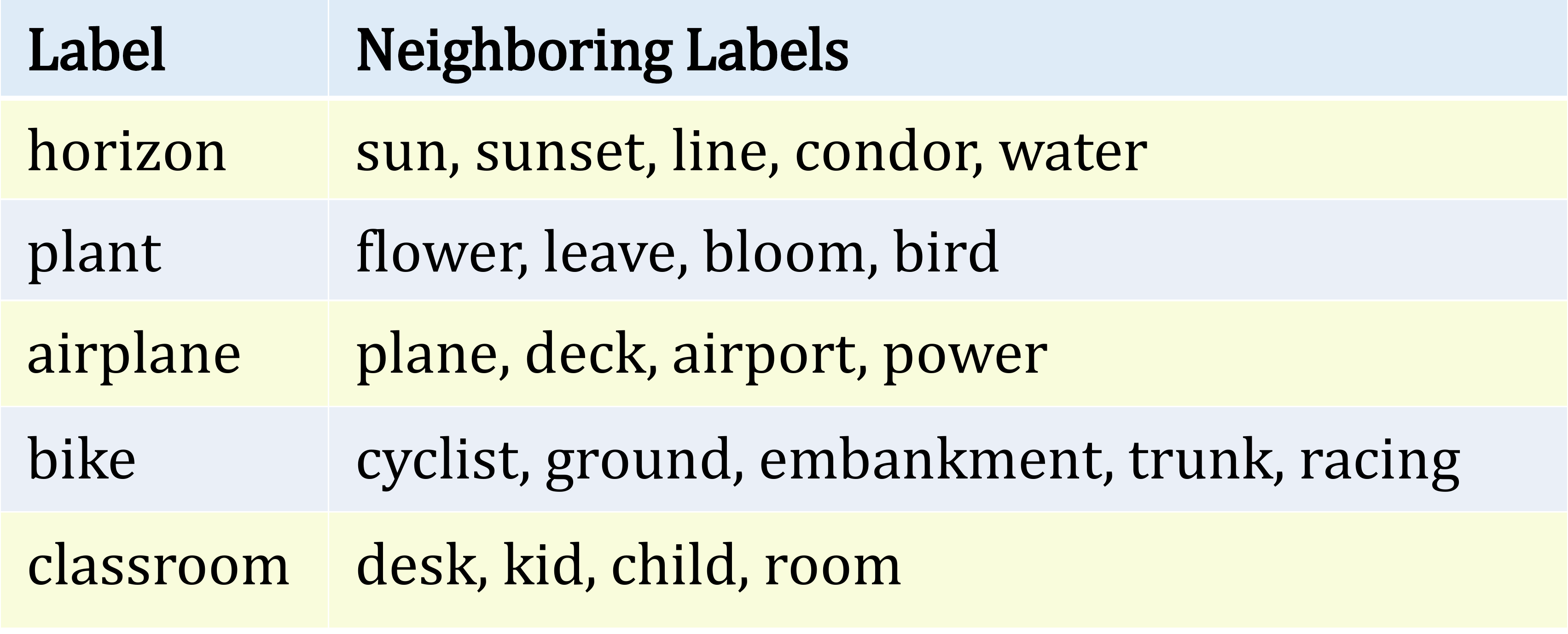}
	\vspace{-2mm}
	\caption{\small Visualization of embedded labels for IAPRTC-12.}
	\vspace{-2mm}
	\label{nearest}
\end{figure}

Figure \ref{fig:LSDR} illustrates and compares the performances of the above methods, in which the horizontal axis denotes the ratio of the latent space dimension ($l$/$m$). From this figure, we see that our C2AE performed favorably against all label embedding methods (with and without DNN introduced) in most cases, which supports our exploitation of nonlinear joint feature and label embedding for multi-label classification. We also see that, with the introduction of DNN architectures for CPLST and FAIE, their DNN versions were not able to achieve comparable performances as ours did. This further verifies the effectiveness of our C2AE in learning deep latent spaces from both feature and label data, and with additional abilities in identifying label co-occurrences.


To further verify the effectiveness of our derived deep latent space, we consider several example labels from \textit{IAPRTC-12}, and list their corresponding neighboring ones in Figure \ref{nearest}. From this figure, we see that the neighboring labels observed in the latent space exhibit highly correlated semantic information. This confirms our C2AE in sufficiently exploiting label dependency during the learning process.

\subsection{Comparisons with DNN-based Approaches}

We further compare our C2AE with recent DNN-based methods for multi-label classification. In addition of a basline method of DNN (as a deep version of binary relevance with the loss function of BCE \cite{nam2014large} and BP-MLL \cite{zhang2006multilabel}), we have (1) WARP \cite{gong2013deep}, which is a CNN network with the WARP loss function, and (2) CNN-RNN \cite{wang2016cnn}, which is a state-of-the-art DNN combining CNN and RNN for multi-label prediction.

The large-scale image annotation dataset of \textit{NUS-WIDE} is applied for evaluation and comparisons. As noted earlier, for fair comparison purposes, we extract 4096-dimensional fc7 features from NUS-WIDE using a pre-trained AlexNet network \cite{krizhevsky2012imagenet} as the feature inputs for C2AE and other methods. And, since existing DNN approaches do not perform dimension reduction from the label space, we fix our dimension reduction ratio $l/m$ as 1. The metrics of per-class and overall precision (C-P and O-P), including the recall scores (C-R and O-R) are considered in accordance with \cite{gong2013deep,wang2016cnn}.

\begin{table}[]
	\centering
	\caption{\small Performance comparisons of DNN-based approaches on NUS-WIDE. Macro-F1 and Micro-F1 are abbreviated as as C-F1 and O-F1, respectively.}
	\footnotesize
	\begin{tabular}{l||lll||lll}
		\hline
		Method   & C-P  & C-R  & C-F1 & O-P  & O-R  & O-F1 \\ \hline
		CNN-WARP & $31.7$ & $35.6$ & $33.5$ & $48.6$ & $60.5$ & $53.9$ \\
		CNN-RNN  & $40.5$ & $30.4$ & $34.7$ & $49.9$ & $61.7$ & $55.2$ \\
		DNN-BCE  & $42.2$ & $23.7$ & $21.7$ & $56.6$ & $67.0$ & $61.4$ \\
		BP-MLL   & $44.5$ & $39.8$ & $38.3$ & $57.3$ & $68.9$ & $62.5$ \\	
		C2AE     & $\mathbf{55.8}$ & $\mathbf{45.3}$ & $\mathbf{48.6}$ & $\mathbf{66.2}$ & $\mathbf{69.1}$ & $\mathbf{67.6}$ \\ \hline
	\end{tabular}
	\label{table:deep}
\end{table}

Table \ref{table:deep} lists and compares the classification performances of different DNN-based methods. It can be seen that DNN-BCE and CNN-WARP did not exhibit abilities in exploiting label co-occurrence information, so they were not able to achieve satisfactory performances. While such capabilities were introduced in BP-MLL and CNN-RNN via linear embedding, our approach still produced promising performances among all DNN methods considered. This supports our use of DNN models in both feature/label embedding and label correlation exploitation.

To make additional remarks on the computation time, our C2AE only takes 10-15 minutes to perform training on \textit{NUS-WIDE} using a titan X GPU, which is much more efficient than training other DNN-based approaches, especially those require the learning of RNN. Nevertheless, our C2AE not only achieves satisfactory classification performance, it is also an efficiently preferable DNN model to consider.

\subsection{Performance Evaluation with Missing Labels}

Finally, we handle the challenging task in which missing labels are presented in the training set. To conduct the experiments, we vary the label missing rate from $10\%$ to $50\%$, while enforcing at least one positive label to be preserved for each instance. Three state-of-the-art approaches are now considered: (1) LEML, (2) Multi-label Learning with Missing Labels (MLML) \cite{wu2014multi}, and (3) ML-MG (i.e., Multi-label Learning with Missing Labels Using a Mixed Graph (ML-PGD) \cite{wu2015ml}).
\begin{figure}[t!]
	\centering
	\includegraphics[width=0.50\textwidth]{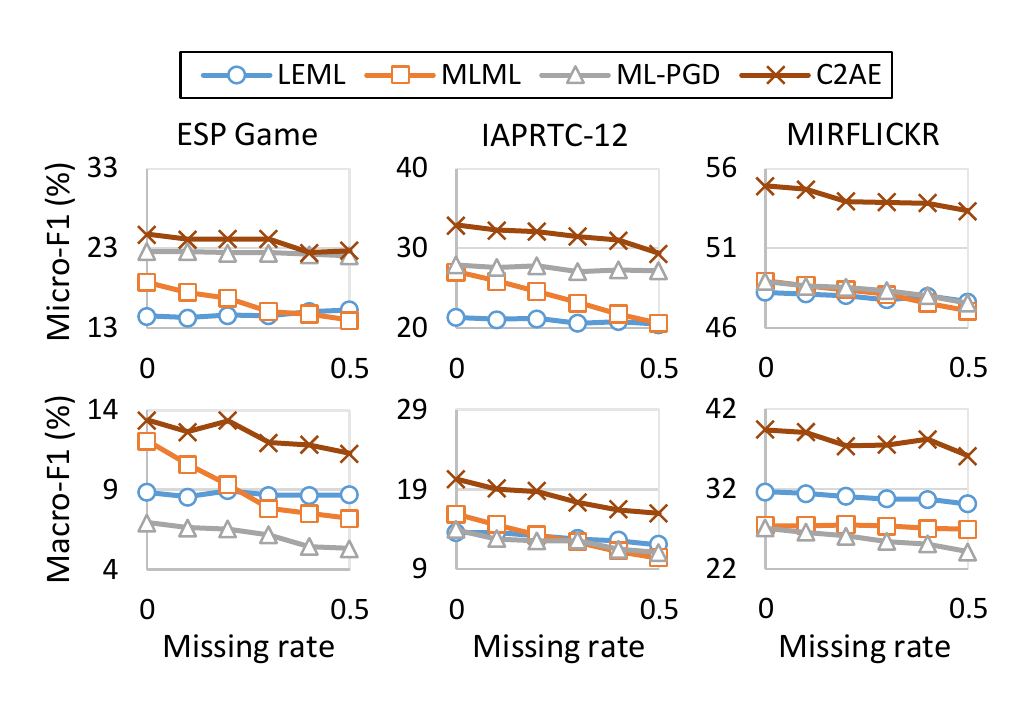}
	\vspace{-2mm}
	\caption{\small Comparisons of Micro-F1 and Macro-F1 with varying label missing rates.}
	\vspace{-2mm}
	\label{fig:missing}
\end{figure}
We show the performance comparisons in Figure \ref{fig:missing}, in which our C2AE consistently and remarkably performed against other approaches. It is worth noting that, existing solutions typically learn linear regressors as their predictors, with additional regularization to handle missing labels. Our C2AE uniquely performs an end-to-end learning with joint feature and label embedding. Its effectiveness for multi-label classification and robustness to missing label problems can be successfully verified by the above experiments.


\section{Conclusion}
We proposed Canonical Correlated Autoencoder (C2AE) for solving the task of multi-label classification. By uniquely integrating DCCA and autoencoder in a unified DNN model, we are able to perform joint feature and label embedding for relating such cross-domain data. With label-correlation sensitive loss functions introduced at the outputs of C2AE, additional ability of exploiting cross-label dependency is further introduced into our learning model. In the experiments, we showed that our C2AE not only performed favorably against baseline and state-of-the-art methods on multiple datasets, we further confirmed that our C2AE can be easily applied for learning tasks with varying amounts of missing labels. Thus, the effectiveness and robustness of our proposed method can be successfully verified.

\section{Acknowledgements}
This work was supported in part by the Ministry of Science and Technology of Taiwan under Grant MOST105-2221-E-001-028-MY2.

\bibliographystyle{aaai}
\bibliography{aaai2017_ref}
\end{document}